\documentclass{article}

\usepackage{arxiv}

\usepackage{newtxtext}
\usepackage{newtxmath}
\newcommand{\tj}[1]{{\fontencoding{T2A}\selectfont #1}}
\usepackage[utf8]{inputenc} % allow utf-8 input
\usepackage[T1]{fontenc}    % support for Cyrillic (T2A) and Latin (T1)
\usepackage[russian,english]{babel} % hyphenation rules for Russian and English
\usepackage{hyperref}       % hyperlinks
\usepackage{url}            % simple URL typesetting
\usepackage{booktabs}       % professional-quality tables
\usepackage{amsfonts}       % blackboard math symbols
\usepackage{nicefrac}       % compact symbols for 1/2, etc.
\usepackage{microtype}      % microtypography
\usepackage{cleveref}       % smart cross-referencing
\usepackage{lipsum}         % Can be removed after putting your text content
\usepackage{graphicx}
\usepackage{natbib}
\usepackage{doi}

\usepackage{xcolor}
\definecolor{jsonstring}{rgb}{0.16,0.45,0.12}
\definecolor{jsonkey}{rgb}{0.55,0.06,0.55}
\definecolor{jsonnumber}{rgb}{0.75,0.15,0.15}

\newif\ifuniqueAffiliation
\uniqueAffiliationfalse  % режим нескольких авторов

\title{Tatarstan Toponyms: A Bilingual Dataset and Hybrid RAG System for Geospatial Question Answering}

\ifuniqueAffiliation
    \author{%
        \href{https://orcid.org/0000-0003-2525-1183}{\includegraphics[scale=0.06]{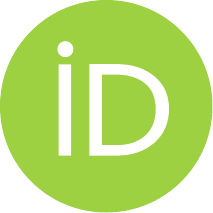}\hspace{1mm}M. K. Arabov}\thanks{Email: \texttt{MKArabov@kpfu.ru}} \\
        Institute of Computational Mathematics and Information Technologies\\
        Kazan Federal University\\
        Kazan, Russia \\
        \texttt{MKArabov@kpfu.ru}
    }
\else
    \usepackage{authblk}
    
    \newbox{\orcid}\sbox{\orcid}{\includegraphics[scale=0.06]{orcid.pdf}}
    \author[1]{%
        \href{https://orcid.org/0000-0003-2525-1183}{\usebox{\orcid}\hspace{1mm}M. K. Arabov}\thanks{\texttt{MKArabov@kpfu.ru}}%
    }
    \author[1]{S.~S.~Khaybullina\thanks{\texttt{khaybulinas@mail.ru}}}
    \author[1]{D.~M.~Naumetova\thanks{\texttt{ndinara03@gmail.com}}}
    \affil[1]{Department of Data Analysis and Programming Technologies, Institute of Computational Mathematics and Information Technologies, Kazan (Volga Region) Federal University, Kazan, Russia}
\fi

\renewcommand{\shorttitle}{Tatarstan Toponyms: Dataset and RAG System}

\hypersetup{
    pdftitle={Tatarstan Toponyms: A Bilingual Dataset and Hybrid RAG System for Geospatial Question Answering},
    pdfsubject={cs.CL, cs.AI},
    pdfauthor={M. K. Arabov, S. S. Khaybullina, D. M. Naumetova},
    pdfkeywords={toponyms, geospatial question answering, RAG, bilingual dataset, Tatarstan, natural language processing},
}

\begin{document}

\maketitle

% ==============================================================================
% ABSTRACT (вставлен из предоставленного текста)
% ==============================================================================
\begin{abstract}
This paper addresses the end-to-end task of automatic geospatial question answering over multilingual toponymic data. An original bilingual (Russian--Tatar) dataset of toponyms of the Republic of Tatarstan is introduced, comprising 9,688 structured records with detailed linguistic, etymological, administrative, and coordinate information (93.1\% of objects are georeferenced). Based on this dataset, a specialized question-answering corpus of approximately 39,000 ``question--context--extractable answer'' triples is constructed, with guaranteed answer localization within the text. To solve the task, an architecture combining two key components is proposed: a hybrid retriever that integrates dense semantic indexing using multilingual-e5-large with a geospatial filter and ranking (KD-trees, haversine distance), and an extractive reader based on fine-tuned transformer models. On 500 test queries, the hybrid search achieves Recall@1~=~0.988, Recall@5~=~1.000, and MRR~=~0.994, statistically significantly outperforming both BM25 and the purely spatial method. Among the tested reader architectures (RuBERT, XLM-RoBERTa-large, T5-RUS), the best answer extraction quality is attained by the multilingual XLM-RoBERTa-large model: EM~=~0.992, F1~=~0.994. A contrasting effect is observed: on raw outputs, RuBERT-based models fail to answer coordinate-related questions (F1~=~0), whereas XLM-RoBERTa-large achieves F1~=~0.984; however, simple post-processing completely eliminates the gaps in numerical values and restores RuBERT accuracy to 100\%. This discrepancy is attributed to tokenization specifics and the composition of pre-training corpora. The created resources (dataset, QA corpus, trained model weights, and web demonstrator) are openly published on the Hugging Face platform. The obtained results can be directly applied in the development of geospatial question-answering services, geocoding systems, and digital humanities projects that benefit from etymological and multilingual place-name data.
\end{abstract}

% ----------------------------------------------------------------------
% KEYWORDS (адаптировано под формат шаблона)
% ----------------------------------------------------------------------

% ==============================================================================
% SECTION 1: INTRODUCTION (вставлен из предоставленного текста)
% ==============================================================================

\section{Introduction}
Geographical names constitute a fundamental component of spatial data infrastructure and play a key role in cartography, navigation, regional governance, and digital humanities research \citep{suleymanov2021_system, galimov2023_dialectologist}. In multilingual regions such as the Republic of Tatarstan, where Russian and Tatar are both official state languages, toponyms function simultaneously in two linguistic codes and accumulate rich etymological, historical, and dialectological information. A notable feature of the local toponymy is the coexistence of several administrative categories of rural settlements: \textit{selo} (\tj{село}) and \textit{derevnya} (\tj{деревня}) are distinct types, the former historically larger and usually possessing a church, while the latter is a smaller settlement. Both are conventionally rendered as ``village'' in English, yet they represent different object types in the source data and are retained as separate classes in our dataset.

Traditional geospatial resources (GeoNames, OpenStreetMap) provide basic coordinate and classification data; however, they lack deep linguistic and etymological details and are not designed for semantic search that accommodates synonymy of geographical terms (e.g., ``\tj{река}''~--~``\tj{приток}'' [river~--~tributary]) and multilingual spelling variants of the same object \citep{rehurek2006_gensim}. At the same time, etymological information, which is present in about 63\% of our records, enables a qualitatively different kind of query: ``Why is Lake Kaban so named?'' or ``What is the origin of the name Sarmanovo?'' Such explanatory question answering is increasingly demanded by educational platforms, digital tourism services, and cultural heritage projects, yet it remains absent from conventional geospatial tools.

In parallel, the past decade has witnessed notable progress in the computational processing of the Tatar language; representative text corpora, morphological analyzers, tokenizers, and semantic annotation tools have been created (see Section~2.2 for a concise overview). Nevertheless, specialized question-answering datasets that account for toponymic specificity have been absent until now. Existing QA collections (SQuAD, SberQuAD, and their Russian-language counterparts) are oriented toward news and encyclopedic texts and do not cover structured geographic information that includes coordinates, object types, and etymological descriptions.

Modern generative models underlying Retrieval-Augmented Generation (RAG) require reliable retrieval components capable of supplying relevant context. In the domain of geographic search, the combination of dense semantic embeddings obtained from multilingual transformers (multilingual E5 \citep{liang2022_e5}, XLM-RoBERTa \citep{conneau2020_xlmr}) with geospatial ranking and filtering is of particular interest. However, the integration of such approaches for multilingual toponymic data, especially coupled with subsequent extractive reading, remains underexplored. The identified gap---the simultaneous absence of both a dataset and an end-to-end question-answering architecture for geographic queries in a bilingual setting---determines the relevance of the present work.

\section{Related Work}
\subsection{Toponymic Data and Resources for the Tatar Language}
International projects such as GeoNames and OpenStreetMap provide open coordinate and attribute information; however, they do not contain the detailed etymological and linguistic data necessary for scholarly onomastics \citep{geonames}. Fundamental academic works on the toponymy of Tatarstan include the hydronym dictionaries and monographs by Garipova \citep{garipova1984_hydronyms1,garipova1990_hydronyms2,garipova1991_hydronymy,garipova2010_toponyms}, Sattarov's comprehensive study of Tatar toponymy \citep{sattarov1998_toponymy}, the two-volume etymological dictionary of the Tatar language by \"{A}khm\"{a}t\"{y}anov \citep{akhmetianov2015_etymology}, the multi-volume Tatar Encyclopedia \citep{tatar_encyclopedia}, and the large dialectological dictionary \citep{bayazitova2009_dialect}. These sources, which have been partially digitized on the ``Toponyms of Tatarstan'' portal, provide the authoritative foundation for the dataset created in the present work (see Section~4.1).

\subsection{Tools for Automatic Processing of the Tatar Language}
For the Tatar language, general-purpose corpora \citep{saykhunov_corpus_tatar, sketch_engine_tatar, ipsan_tat_monocorpus}, morphological analysis systems \citep{gilmullin2017_morphological}, tokenizers \citep{arabov2026_tatartokenizers}, as well as methods for semantic annotation based on knowledge graphs \citep{mukhamedshin2025_semantic} and machine learning \citep{mindubaev2024_semantic_relation}, have been developed. Software solutions for constructing vector representations of words and documents have been created \citep{arabov2026_tatar2vec, gafarov2025_explainable}. All of the listed resources are consolidated on the Hugging Face platform within the TatarNLPWorld organization, ensuring reproducibility of experiments \citep{tatarnlpworld_hf}.

\subsection{Semantic and Geospatial Search}
Classical lexical methods such as BM25 \citep{rehurek2006_gensim} struggle with cross-lingual and synonymic queries. The development of Transformer architectures has led to the emergence of dense retrieval models. The E5 family \citep{liang2022_e5} is trained with a contrastive loss function on multilingual collections and is capable of encoding documents and queries into a unified semantic space. The multilingual-e5-large model demonstrates high effectiveness for low-resource and multilingual scenarios \citep{conneau2020_xlmr, schweter2020_berturk}. In parallel, geoinformatics employs spatial indices (KD-trees \citep{bentley1975_kdtree}, R-trees) and metrics that account for the Earth's curvature (haversine distance \citep{sinnott1984_haversine}). Works \citep{galimov2023_dialectologist, burnashev2025_geolinguistic} proposed geolinguistic systems combining fuzzy logic with spatial data for dialect analysis; however, they did not utilize dense embeddings. Hybrid approaches that combine textual and geographic relevance in a weighted manner have been successfully applied in recommendation services and point-of-interest search, but such solutions have not been previously investigated for multilingual toponyms with detailed etymological attribution.

\subsection{Question-Answering Systems and Extractive Reading}
For the extractive QA task, SQuAD \citep{rajpurkar2016_squad} and its multilingual versions have become the de facto standard datasets. Russian-language counterparts, such as SberQuAD \citep{efimov2020_sberquad}, are oriented primarily toward news and encyclopedic texts and do not contain structured geographic data with coordinates. Broader diagnostic benchmarks for the Russian language, in particular Russian SuperGLUE \citep{fenogenova2021_rusuperglue}, enable the evaluation of multiple aspects of language understanding but also do not include tasks requiring the extraction of numerical coordinates and etymological information. Among models, BERT-based architectures lead the field: RuBERT \citep{devlin2019_bert} (further pre-trained on Russian texts) and the multilingual XLM-RoBERTa \citep{conneau2020_xlmr}, trained on one hundred languages. The latter has demonstrated excellent results in recognizing nested named entities, including geographical ones, as confirmed, for example, within the Russian-language competition RuNNE-2022 \citep{artemova2022_runne}. Generative T5 models \citep{raffel2020_t5} are also applied, but typically in the answer generation setting. A significant limitation of monolingual models, revealed in recent experiments \citep{choure2022_hindi_ner}, is their inability to process numerical coordinates due to WordPiece tokenization specifics and a shortage of relevant examples in training sets. Multilingual models employing SentencePiece, by contrast, successfully extract numerical sequences. To date, no specialized QA benchmarks combining structured geographic information with etymological and coordinate components have been proposed.

Thus, the literature review attests to the existence of disparate components (language resources for Tatar, dense retrieval models, geospatial filtering methods, extractive reading models); however, their integration into a unified system capable of finding the required toponym and extracting the precise answer for an arbitrary geographic query in Russian or Tatar is lacking. The present work is aimed at filling this gap.

\section{Methodology}
\subsection{Hybrid Retrieval: Architecture and Method}
The purpose of the retrieval component is to take a textual query \(q\), optionally accompanied by a geographic point \((\textit{lat}_q, \textit{lon}_q)\) and a radius \(R\), and produce a ranked list of the \(k\) most relevant toponyms from a set of documents \(D\), each of which is equipped with coordinates. The architecture comprises two parallel indexing stages---semantic and spatial---which interact during query processing in a hybrid ranking procedure.

\paragraph{Semantic Indexing.}
To represent the content portion of a document, the multilingual encoder model \texttt{intfloat/multilingual-e5-large} \citep{liang2022_e5} is employed, which transforms the contextual field \texttt{context} into a 1024-dimensional dense vector. The model was chosen as it demonstrates state-of-the-art results in cross-lingual search and supports both Russian and Tatar languages. All document embeddings are normalized to unit \(L_2\)-norm, allowing cosine proximity to be assessed via the dot product. Computation is performed in batches of 32 documents using a GPU. For efficient nearest-neighbor search, a flat inner product index (\texttt{IndexFlatIP}) is constructed using the FAISS library \citep{johnson2019_faiss}; when a GPU accelerator is available, the index is placed in GPU memory. Query processing reduces to encoding its text with the same model, \(L_2\)-normalizing the query vector, and extracting the top-\(k\) documents with the maximum dot product. The resulting quantity
\[
\textit{sem\_score}(i) = \langle e_q, e_i \rangle
\]
is interpreted as the semantic relevance of document \(i\).

\paragraph{Geospatial Filter and Ranking.}
Spatial selection is implemented in two passes. In the first stage, a bounding box is constructed around the query point:
\[
\Delta\textit{lat} = R / 111320, \quad \Delta\textit{lon} = R / (111320 \cdot \cos(\phi)),
\]
where \(\phi\) is the latitude in radians and \(R\) is the search radius. Objects whose coordinates fall within this region are declared candidates. Then, for each candidate, the exact haversine distance \citep{sinnott1984_haversine} is computed, and those objects whose distance exceeds \(R\) are filtered out. The spatial score of the remaining candidates is determined by an exponentially decaying function of distance:

\[
geo\_score(i) = e^{-\frac{d_i}{R}}
\]

To accelerate the filtering stage, a KD-tree \citep{bentley1975_kdtree} is built over the coordinates of all documents during corpus preparation, enabling logarithmic-time retrieval of the subset of objects within a given neighborhood.

\paragraph{Combined Relevance.}
The final relevance is formed as a weighted sum of normalized semantic and spatial scores. Semantic scores are normalized using min-max scaling among the candidates that passed the geofilter:
\[
\textit{sem\_norm}(i) = \frac{\textit{sem\_score}(i) - \min_{\textit{sem}}}{\max_{\textit{sem}} - \min_{\textit{sem}}},
\]
and in degenerate cases (when the difference is close to zero), all values are set to 1. Spatial scores are normalized by dividing by the maximum value among the candidates:
\[
\textit{geo\_norm}(i) = \frac{\textit{geo\_score}(i)}{\max_{\textit{geo}}}.
\]
The combined score is computed as
\[
\textit{score}(i) = \alpha \cdot \textit{sem\_norm}(i) + (1 - \alpha) \cdot \textit{geo\_norm}(i),
\]
where \(\alpha \in [0,1]\) is a hyperparameter governing the contribution of the semantic component. Objects are ranked by descending \(\textit{score}(i)\), and the top-\(k\) are returned. The value of \(\alpha\) is tuned empirically on a validation set (200 queries) via grid search over \(\{0.1, 0.3, 0.5, 0.7, 0.9\}\) by maximizing the average Recall@5. In the experiments, \(\alpha = 0.1\) was found to be optimal, with a fixed radius of \(R = 50\) km, chosen based on the typical scale of local search within Tatarstan.

\subsection{Extractive Reading: QA Corpus Generation and Model Training}
For the extractive reading component, the original toponym dataset is transformed into a set of extractive question-answer pairs with precise answer spans. The procedure, described in full in Section~4.2, guarantees that every answer appears verbatim within the context and covers all major information categories (object type, location, etymology, coordinates, etc.). In total, 38,696 pairs were generated and split into training (90\,\%) and validation (10\,\%) sets, stratified by question type.

\paragraph{Models and Training.}
Two groups of transformer architectures were fine-tuned for the reading task: monolingual Russian-language RuBERT models (base and large versions, initialized from SberQuAD checkpoints \citep{efimov2020_sberquad, devlin2019_bert}) and the multilingual XLM-RoBERTa-large model \citep{conneau2020_xlmr} pre-trained on SQuAD 2.0. The generative model T5-RUS \citep{raffel2020_t5} was used without fine-tuning as a reference point due to technical incompatibilities. All models were trained on the generated corpus for three epochs with the AdamW optimizer, a learning rate of \(3 \times 10^{-5}\), batch size 4, and linear warm-up over 500 steps. Extractive models predict start and end positions of the answer span; T5-RUS was prompted in the format ``question: \ldots\ context: \ldots\ \(\to\) answer''. A simple heuristic baseline that extracts the answer based on question keywords and field prefixes was also implemented.

The trained models, together with the QA corpus, are published on Hugging Face \citep{tatarnlpworld_toponyms_qa} to ensure reproducibility.

\section{Data}
\subsection{Source Dataset of Tatarstan Toponyms}
The empirical foundation of the study is a specialized bilingual (Russian--Tatar) dataset of Tatarstan toponyms, collected, structured, and published by the authors in open access on the Hugging Face platform \citep{tatarnlpworld_toponyms}. The necessity of this resource stems from the absence of machine-readable datasets that simultaneously contain coordinate, linguistic, and etymological information about the region's geographical objects. The dataset draws on fundamental academic works on the toponymy of Tatarstan: the hydronym dictionaries and monographs by Garipova \citep{garipova1984_hydronyms1,garipova1990_hydronyms2,garipova1991_hydronymy,garipova2010_toponyms}, Sattarov's comprehensive study \citep{sattarov1998_toponymy}, the two-volume etymological dictionary by \"{A}khm\"{a}t\"{y}anov \citep{akhmetianov2015_etymology}, the multi-volume Tatar Encyclopedia \citep{tatar_encyclopedia}, the large dialectological dictionary \citep{bayazitova2009_dialect}, and materials from the digital portal ``Toponyms of Tatarstan'' (toponym.antat.ru). Reliance on peer-reviewed academic sources guarantees high reliability and completeness of the data.

The total volume of the dataset amounts to 9,688 records, each corresponding to a single geographical object within the Republic of Tatarstan and adjacent regions of compact Tatar settlement. The record structure includes the following fields: a unique identifier, source URL, toponym type (toponym or microtoponym), toponym subtype (oikonym, hydronym, oronym, or no type), geographical object (76 unique categories: village, selo,\footnote{In Russian administrative classification, ``village'' (\tj{деревня}) and ``selo'' (\tj{село}) are distinct types of rural settlements; the latter is historically larger and usually possessed a church. Both are translated as ``village'' in English but are kept separate in the dataset to preserve the original distinction.} river, lake, field, mountain, meadow, river tributary, glade, etc.), name in Russian, name in Tatar, federal subject, physiographic details, geographical location, name etymology, bibliographic sources, latitude and longitude coordinates (where available), and a map availability flag. Etymological information, of particular value for onomastic and historical-geographic research, is present in approximately 63\,\% of records.

A key characteristic of the dataset is the presence of coordinate referencing: 9,023 records (93.1\,\%) are furnished with latitude and longitude values, enabling their use in geospatial analysis and distance-based filtering. The remaining 665 records (6.9\,\%) without coordinates were excluded from the hybrid search experiments but were still used for generating QA pairs that do not require spatial information (etymology, sources).

The distribution of records by toponym subtype is presented in Table~\ref{tab:toponym_subtype}. The predominant share consists of oikonyms---names of populated places (villages, selos, settlements), which mirrors the settlement structure of the region and the thematic coverage of the sources used. Hydronyms (names of water bodies: rivers, lakes, streams) and oronyms (names of landforms: mountains, hills, ravines) together constitute about one quarter of the corpus. The ``no type'' category unites records for which the toponym subtype was not explicitly indicated in the sources, although they retain information about the geographical object and etymology.

\begin{table}[htbp]
\centering
\begin{tabular}{lcc}
\toprule
\textbf{Toponym Subtype} & \textbf{Number of Records} & \textbf{Share, \%} \\
\midrule
Oikonym & 7000 & 72.2 \\
Hydronym & 1500 & 15.5 \\
Oronym & 800 & 8.3 \\
No type & 388 & 4.0 \\
\midrule
\textbf{Total} & \textbf{9688} & \textbf{100.0} \\
\end{tabular}
\caption{Distribution of dataset records by toponym subtype}
\label{tab:toponym_subtype}
\end{table}

Granularity by geographical object type (Table~\ref{tab:geo_objects}) reveals the most frequent categories. Populated places dominate: villages (36.1\,\%) and selos (20.6\,\%), which corresponds to the share of oikonyms in Table~\ref{tab:toponym_subtype}. Among natural objects, the most represented are meadows (8.3\,\%), rivers (7.2\,\%), fields (5.2\,\%), and mountains (4.1\,\%). About 10.7\,\% of records belong to other, less frequent categories, such as lakes, river tributaries, glades, springs, natural tracts, swamps, and ravines. The presence of 76 unique geographical object types ensures high entity diversity and allows testing search algorithms under a heterogeneous taxonomy.

\begin{table}[htbp]
\centering
\begin{tabular}{lcc}
\toprule
\textbf{Geographical Object} & \textbf{Share, \%} \\
\midrule
Village & 36.1 \\
Selo & 20.6 \\
Meadow & 8.3 \\
River & 7.2 \\
Field & 5.2 \\
Mountain & 4.1 \\
Lake & 3.1 \\
River tributary & 2.6 \\
Glade & 2.1 \\
Other & 10.7 \\
\midrule
\textbf{Total} & \textbf{100.0} \\
\end{tabular}
\caption{Top-10 most frequent geographical objects in the dataset}
\label{tab:geo_objects}
\end{table}

The regional distribution reflects the geographic focus of the dataset: approximately 93\,\% of records concern objects within the Republic of Tatarstan, about 4\,\% belong to Tyumen Oblast (areas of compact settlement of Siberian Tatars), and the remaining 3\,\% are distributed among adjacent subjects of the Russian Federation (Bashkortostan, Ulyanovsk, Samara, Orenburg Oblasts, and others).

To facilitate semantic search, all textual fields of each record, excluding coordinates, were merged into a single contextual field \texttt{context} using English-language key prefixes. The context format is as follows:

\begin{quote}
\ttfamily Name (rus): <name\_rus> | Name (tat): <name\_tat> | Type: <toponym\_type> | Subtype: <toponym\_subtype> | Object: <geographical\_object> | Etymology: <etymology> | Details: <physiographic\_details> | Location: <geographical\_location> | Sources: <bibliographic\_sources>
\end{quote}

Fields containing no information were excluded from the context. The choice of English-language prefixes is motivated by the use of the multilingual model \texttt{multilingual-e5-large}, for which English-language keys provide more robust cross-lingual matching of semantically similar fields. This procedure yielded 9,023 documents with non-empty contextual fields and coordinate referencing, which formed the corpus for indexing and testing the retrieval component.

\subsection{Generation of the Question-Answering Corpus}
Based on the structured toponym dataset, a synthetic question-answering corpus was automatically generated for training and evaluating extractive QA models. The central requirement was that every answer be guaranteed to appear verbatim within the context string, enabling precise character-level annotation of the answer span.

For each information type in a source record (name, object type, location, etymology, coordinates, region, physiographic characteristics, sources), the corresponding field was transformed into a string with a Russian-language prefix that unambiguously identifies the category: \texttt{Name (Rus):}, \texttt{Name (Tat):}, \texttt{Object:}, \texttt{Etymology:}, \texttt{Location:}, \texttt{Coordinates:}, \texttt{Region:}, \texttt{Physiographic Details:}, \texttt{Sources:}. All prefixed strings were concatenated via the ``\texttt{|}'' separator into a single context. If the resulting string exceeded 2048 characters, each field was proportionally truncated while preserving its prefix. This design guarantees that any potential answer appears together with its identifying prefix, so the answer start position can be unambiguously computed as the sum of the lengths of the preceding fields and separators.

Russian-language question templates with a \texttt{\{name\}} placeholder were developed for seven information categories; the complete list is available in the dataset card on Hugging Face.

During generation, one template was randomly selected for each record, and the Russian name (or, if unavailable, the Tatar name) was inserted into the placeholder. The answer was taken directly from the corresponding field (for coordinates, a comma-separated latitude--longitude string). The answer start position was computed algorithmically as described above. Each resulting pair thus carries exact positional labels required for extractive model training.

The maximum number of QA pairs per record was capped at ten to avoid dominance of objects with many populated fields and to ensure uniform coverage. The total volume of the generated corpus is 38,696 pairs. Their distribution by question type is shown in Table~\ref{tab:qa_distribution}.

\begin{table}[htbp]
\centering
\begin{tabular}{lcc}
\toprule
\textbf{Question Type} & \textbf{Number of Pairs} & \textbf{Share, \%} \\
\midrule
Coordinates & 12,344 & 31.9 \\
Object type & 8,032 & 20.8 \\
Location & 5,724 & 14.8 \\
Etymology & 5,032 & 13.0 \\
Region & 3,868 & 10.0 \\
Sources & 3,052 & 7.9 \\
Physiographic characteristics & 644 & 1.6 \\
\midrule
\textbf{Total} & \textbf{38,696} & \textbf{100} \\
\end{tabular}
\caption{Distribution of the synthetic QA corpus by question type}
\label{tab:qa_distribution}
\end{table}

Coordinate questions form the largest share (31.9\,\%), reflecting the primary importance of spatial information. Questions about object type (20.8\,\%) and location (14.8\,\%) together account for more than one third of the corpus. Etymological questions (13.0\,\%) exploit the dataset's unique feature---the availability of name-origin data. The smallest category, physiographic characteristics (1.6\,\%), corresponds to the fragmentary population of that field in the source records.

The average context length is 1250 characters; answers vary from 2 to 150 characters. Coordinate answers are the shortest (around 20 characters on average), while answers containing bibliographic sources can reach up to 500 characters. An illustrative span-annotated example is:

\begin{quote}\small\ttfamily
\{\\
~~\textcolor{jsonkey}{"id"}: \textcolor{jsonstring}{"1530\_coordinates\_0"},\\
~~\textcolor{jsonkey}{"context"}: \textcolor{jsonstring}{"Name (rus): Rantamak | Name (tat): Rantamak | Object: Selo | Etymology: The toponym originates from the oikonym ``Rangazar-Tamak''. | Location: Located on the Melya River, 21~km east of the village of Sarmanovo. | Sources: R.G.~\"{A}khm\"{a}t\"{y}anov, Etymological Dictionary of the Tatar Language... | Coordinates: 55.205461, 52.881862"},\\
~~\textcolor{jsonkey}{"question"}: \textcolor{jsonstring}{"What are the coordinates of Rantamak?"},\\
~~\textcolor{jsonkey}{"answers"}: [\{\textcolor{jsonkey}{"text"}: \textcolor{jsonstring}{"55.205461, 52.881862"}, 
                           \textcolor{jsonkey}{"answer\_start"}: \textcolor{jsonnumber}{312}\}]\\
\}
\end{quote}

The corpus was randomly split into training (90\,\%, 34,826 pairs) and validation (10\,\%, 3,870 pairs) sets while preserving the stratification by question type. Both parts, together with the source toponym dataset, are openly published on the Hugging Face platform \citep{tatarnlpworld_toponyms, tatarnlpworld_toponyms_qa} under the CC BY-SA 4.0 license, and are available for scholarly and applied use.

\section{Experimental Study}
The experimental validation of the proposed question-answering system was conducted in two stages, corresponding to its architectural components. In the first stage, the effectiveness of the hybrid retrieval module, combining semantic indexing with geospatial filtering, was evaluated; in the second, the ability of the extractive reading component to accurately localize the answer within the context provided by the retrieval mechanism was assessed. Such a two-part evaluation protocol made it possible, on the one hand, to characterize each subsystem in isolation, and on the other, to demonstrate their compatibility within a unified pipeline. All experiments were carried out on the corpus of 9,023 coordinate-referenced documents described above and on the synthetic question-answering set of 38,696 examples generated according to the procedure in Section~4.2.

To evaluate the retrieval component, an independent test set was prepared by randomly selecting 500 records from the source dataset and generating natural-language queries for them using five templates (``What is \{name\}?'', ``Where is \{name\}?'', ``Tell about \{name\}'', etc.), in which the object name was inserted either in Russian or in Tatar with probabilities of 0.7 and 0.3, respectively. This approach guaranteed that for each query there exists exactly one relevant document---the source record from which the query was generated. Additionally, a validation set of 200 queries, disjoint from the test set, was allocated; it was used exclusively for tuning the hybrid search hyperparameters---the weighting coefficient \(\alpha\) and the spatial filter radius \(R\).

Retrieval quality was measured by two main metrics: Recall@\(k\) (\(k = 1, 3, 5\)) and Mean Reciprocal Rank (MRR). The former indicates the proportion of queries for which the relevant document appears among the top \(k\) retrieval results; the latter averages the reciprocal of the rank of the first relevant document and is thus sensitive to early hits. To obtain statistically sound conclusions, all metrics were accompanied by 95\% confidence intervals constructed via bootstrap with 1,000 resamples with replacement. This enabled not only comparison of average values but also assessment of the significance of differences between methods.

Four ranking strategies were compared: classical lexical BM25 search, purely spatial search (top-\(k\) nearest by haversine distance), semantic search based on \texttt{multilingual-e5-large} with a FAISS index, and the proposed hybrid method. The hybrid search was performed with a fixed radius of \(R = 50\) km and \(\alpha = 0.1\), selected on the validation set as yielding the maximum \(\text{Recall@5} = 1.0\). The comparison results are summarized in Table~\ref{tab:retrieval_comparison}.

\begin{table}[htbp]
\centering
\small
\begin{tabular}{lcccc}
\toprule
\textbf{Method} & \textbf{Recall@1} & \textbf{Recall@3} & \textbf{Recall@5} & \textbf{MRR} \\
\midrule
BM25 & 0.438 & 0.574 & 0.618 & 0.508 \\
 & {[0.394, 0.484]} & {[0.528, 0.620]} & {[0.572, 0.662]} & {[0.469, 0.545]} \\
Spatial only & 0.536 & 0.708 & 0.796 & 0.634 \\
 & {[0.492, 0.576]} & {[0.668, 0.748]} & {[0.760, 0.830]} & {[0.598, 0.673]} \\
Semantic only & 0.774 & 0.904 & 0.940 & 0.840 \\
 & {[0.736, 0.810]} & {[0.878, 0.928]} & {[0.918, 0.960]} & {[0.813, 0.866]} \\
Hybrid ($\alpha = 0.1$, $R = 50$ km) & 0.988 & 1.000 & 1.000 & 0.994 \\
 & {[0.978, 0.996]} & {[1.000, 1.000]} & {[1.000, 1.000]} & {[0.988, 0.998]} \\
\end{tabular}
\caption{Comparison of search methods (95\% bootstrap confidence intervals)}
\label{tab:retrieval_comparison}
\end{table}

The figures in Table~\ref{tab:retrieval_comparison} indicate a substantial superiority of the hybrid scheme. First of all, the value \(\text{Recall@5} = 1.000\) is striking, meaning that for each of the 500 test queries, the relevant object is guaranteed to be present among the first five results. Such high recall practically eliminates the risk of losing the necessary information before the reading stage, which is a critical requirement for RAG systems. The \(\text{Recall@1} = 0.988\) score indicates that only in six cases out of five hundred did the target document fail to occupy the first position; moreover, the lower bound of the confidence interval (0.978) lies substantially above the upper bounds of \(\text{Recall@1}\) for all competing methods. This arrangement of intervals confirms the statistical significance of the hybrid's superiority over the alternatives.

Semantic search by itself demonstrates fairly high quality (\(\text{Recall@1} = 0.774\)), confirming the ability of the \texttt{multilingual-e5-large} model to adequately encode multilingual toponymic contexts and capture semantic proximity between different name variants. However, the absence of spatial filtering allows objects with similar names but located far apart to appear in top positions, which reduces accuracy by approximately 0.2 compared to the hybrid. Spatial search, which ignores the query text, shows \(\text{Recall@1} = 0.536\). This is expected: within a 50 km radius, several objects are often found, and the geographically nearest one does not always coincide with the sought one. Nevertheless, even this simple strategy outperforms BM25 (\(\text{Recall@1} = 0.438\)), underscoring the fundamental role of geographic proximity for toponymic queries. BM25's most modest results are explained not only by the lexical gap between Tatar and Russian names but also by the synonymy of geographical terms: a classical inverted index is unable to equate ``selo'' and ``village'', ``river'' and ``tributary''.

\begin{figure}[htbp]
\centering
\includegraphics[width=0.7\linewidth]{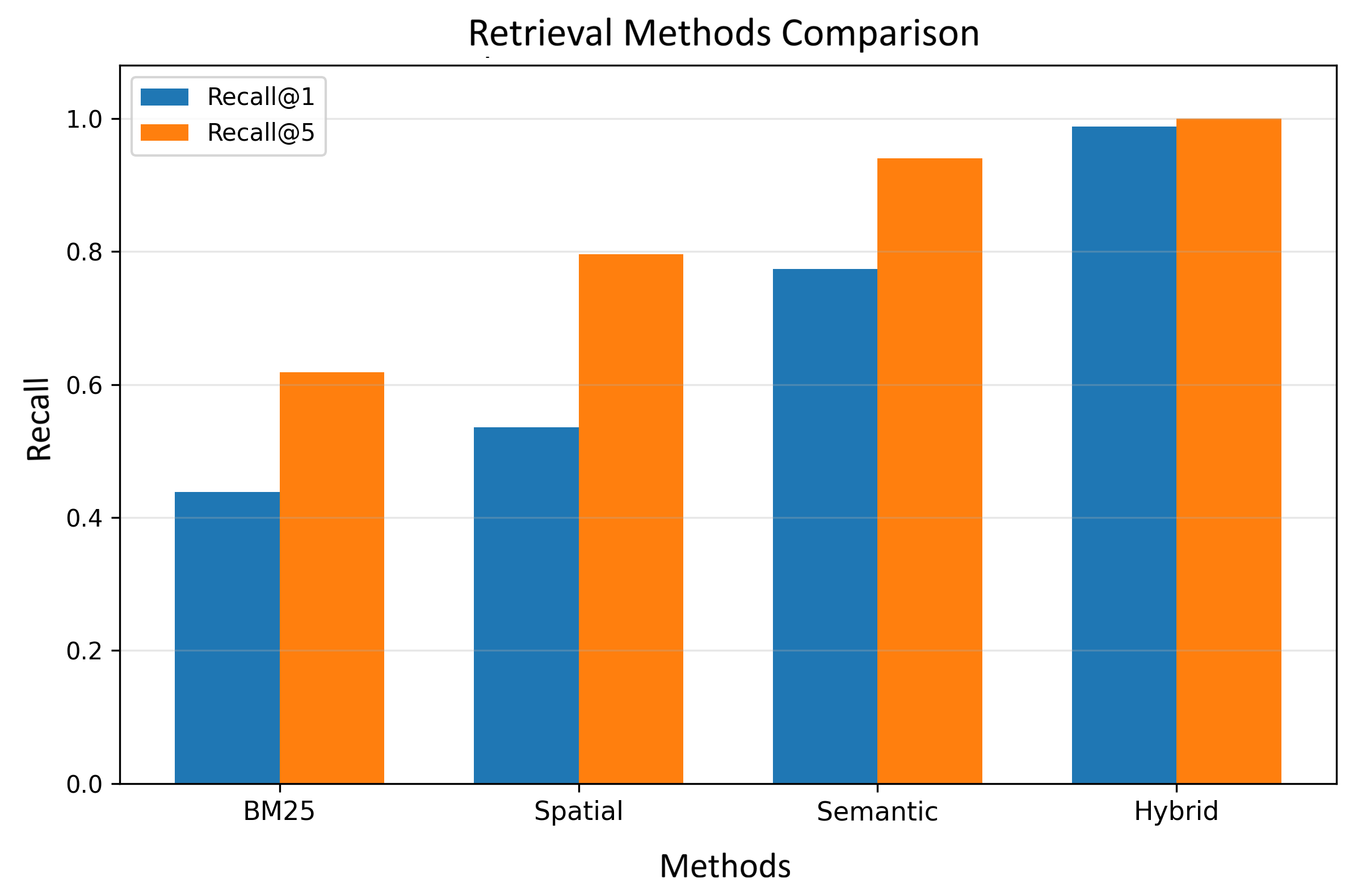}
\caption{Comparison of Recall@1 and Recall@5 with 95\% confidence intervals for BM25, purely spatial, semantic, and hybrid methods.}
\label{fig:retrieval_recall}
\end{figure}

To visually consolidate these comparative findings, Figure~\ref{fig:retrieval_recall} presents a bar chart of Recall@1 and Recall@5 for each method, supplemented with 95\% confidence intervals. This graphical representation not only confirms the numerical superiority of the hybrid method but also reveals a striking pattern: the hybrid method exhibits minimal variability in its error bars, indicating high stability and reproducibility of performance across different query instances. In contrast, the spatial and lexical methods show substantially wider confidence intervals, reflecting greater variance in their retrieval effectiveness. The clear separation between the hybrid confidence intervals and those of all competing methods provides visual confirmation of the statistical significance established earlier.

While the aggregate metrics presented in Table~\ref{tab:retrieval_comparison} and Figure~\ref{fig:retrieval_recall} unequivocally establish the superiority of the hybrid approach over its constituents, the structure of the dataset allows for a more granular analysis. To understand whether the system performs uniformly across different types of geographical features, we next dissected the retrieval results by toponym category. This is important because toponyms vary greatly in their spatial density and linguistic representation. For instance, large populated places (oikonyms) appear frequently in the dataset with rich contexts, while microtoponyms such as meadows or springs have fewer surrounding landmarks and often rely on precise coordinate information. We report the Recall@1 breakdown directly: for toponyms (349 queries) it is 0.986, and for microtoponyms (151 queries) it is 0.993.

\begin{figure}[htbp]
\centering
\includegraphics[width=0.7\linewidth]{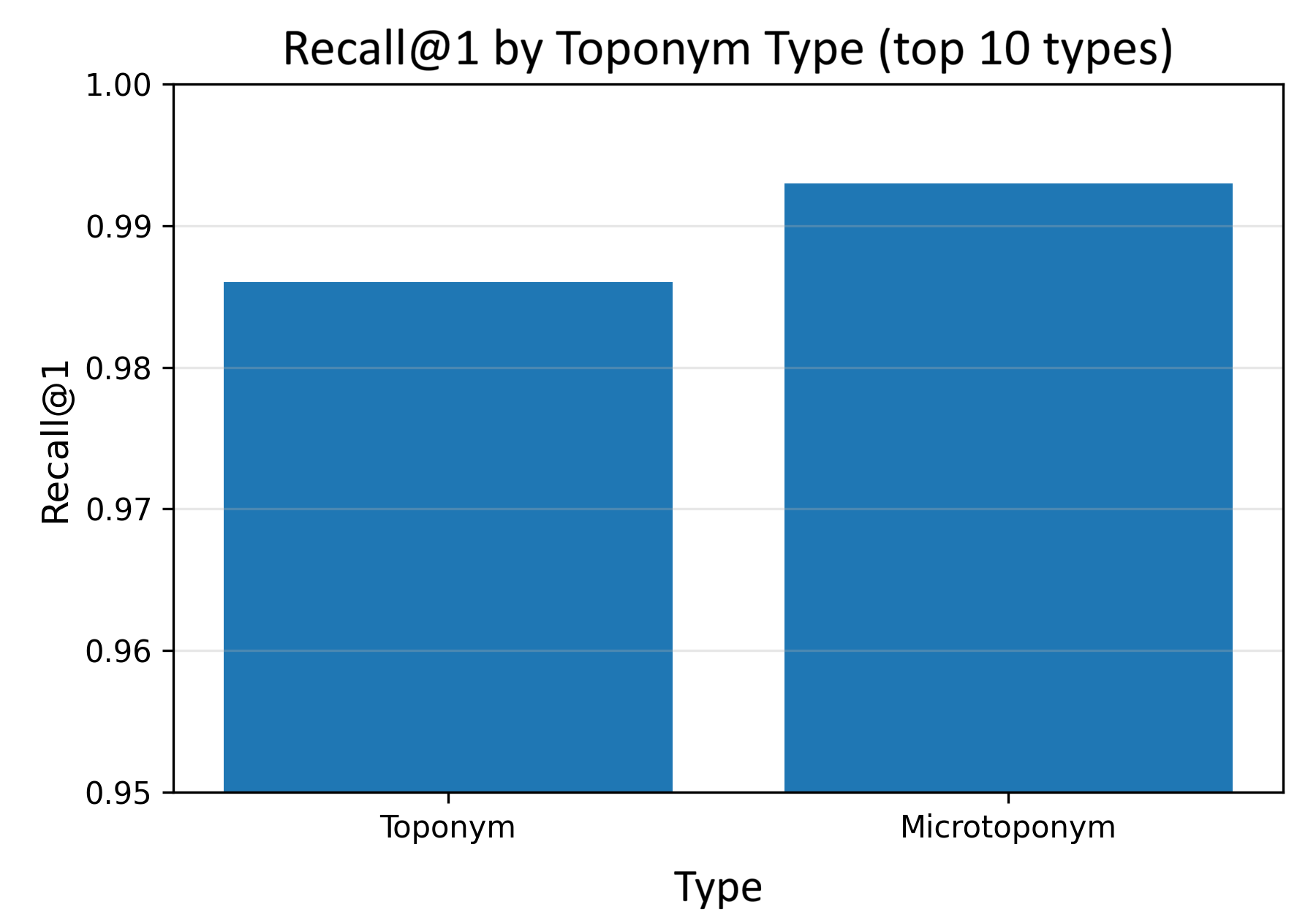}
\caption{Recall@1 of the hybrid method for the ``Toponym'' and ``Microtoponym'' categories.}
\label{fig:toponym_type_recall}
\end{figure}

Both categories demonstrate near-ceiling values; however, microtoponyms show a slightly higher result. This difference is explained by their local nature: for small objects, the density of toponyms within a 50 km radius is generally lower, meaning the spatial filter leaves fewer candidates, and the semantic component finds it easier to identify the sole correct one. In contrast, for larger toponyms, the higher density of historical records and alternative names within the same radius increases the likelihood of semantic confusion, despite the overall high performance. This pattern suggests that the hybrid method scales gracefully to objects of varying granularity, maintaining near-perfect recall even in densely populated regions. Figure~\ref{fig:toponym_type_recall} visualizes this comparison, with the bar chart clearly illustrating the slight but systematic superiority of microtoponym retrieval.

Having established the near-perfect performance across both toponym categories, it is equally instructive to examine the rare instances where the retrieval system falls short. These borderline cases offer critical insights into the system's limitations and, more importantly, reveal concrete pathways for future data-centric improvements. Understanding the nature of these failures allows us to separate algorithmic deficiencies from data quality issues.

Despite the impressive aggregate metrics, the hybrid retrieval made six errors in which the relevant document did not appear in the first position. Manual analysis of these cases, undertaken to identify root causes, showed that all of them are related not to shortcomings of the combination algorithm but to the quality of the data themselves. In two cases, complete namesakes were present within a 50 km radius---two different objects with identical names (e.g., two natural tracts named ``Krasnaya Gorka''). Their contexts proved practically identical, and the semantic component could not give preference to one over the other. In three cases, the coordinates of the target object were offset by 10--15 kilometres from the true position, due to which it did not fall within the search zone and was eliminated at the geofiltering stage. In one further case, the \texttt{context} field contained a minimum of information (only name and type), which yielded a predictably low semantic score and allowed another, more informative candidate to take the lead. Consequently, further improvement of the retrieval component lies primarily in enhancing data quality: verifying coordinates through cross-validation with OpenStreetMap or satellite imagery, enriching contextual fields, and developing a component for resolving toponymic homonymy.

The retrieval analysis confirms that the hybrid algorithm is highly robust and achieves near-ideal recall, with its remaining errors attributable to data quality rather than algorithmic shortcomings. With the search module validated, the next logical step is to evaluate the downstream component responsible for extracting precise answers from the retrieved contexts.

Turning to the second stage, extractive reading, we note that the task here was to precisely locate the answer within the context retrieved by the retrieval module. Three architectures, representing both monolingual and multilingual approaches, were selected for training and evaluation: RuBERT base, RuBERT large, and XLM-RoBERTa large, each fine-tuned on the synthetic QA corpus of 34,826 pairs. The generative model T5-RUS, unfortunately, could not be trained due to tokenizer version incompatibility and was used only in its base pre-trained variant as an additional reference point. Additionally, as a simple baseline, a heuristic rule was implemented that extracts the answer based on question keywords and the corresponding prefixes in the context. The training hyperparameters, common to all models, are presented in Table~\ref{tab:qa_hyperparams}.

\begin{table}[htbp]
\centering
\begin{tabular}{lc}
\toprule
\textbf{Parameter} & \textbf{Value} \\
\midrule
Maximum sequence length & 384 tokens \\
Stride & 128 tokens \\
Training batch size & 4 \\
Evaluation batch size & 8 \\
Learning rate & \(3 \times 10^{-5}\) \\
Number of epochs & 3 \\
Weight decay & 0.01 \\
Warm-up steps & 500 \\
Optimizer & AdamW \\
\end{tabular}
\caption{Hyperparameters for QA model fine-tuning}
\label{tab:qa_hyperparams}
\end{table}

Following the fine-tuning protocol outlined in Table~\ref{tab:qa_hyperparams}, all models were evaluated on the held-out validation set of 3,870 QA pairs. The results, summarized in Table~\ref{tab:qa_results}, revealed a critical nuance regarding tokenizer-induced artifacts that necessitates a careful distinction between raw and normalized performance metrics. It turned out that the RuBERT models, operating with the WordPiece tokenizer, tend to insert extraneous spaces inside numerical coordinates (``55. 175195'' instead of ``55.175195'') and hyphenated constructions (``north - west''). This is a purely tokenization artifact unrelated to the semantic capability of the model. By applying elementary post-processing (removal of breaks within floating-point numbers and unification of hyphens and brackets), we ensured that RuBERT answers became identical to the reference ones. Table~\ref{tab:qa_results} presents both raw Exact Match and F1 metrics and the values obtained after normalization.

\begin{table}[htbp]
\centering
\small
\begin{tabular}{lccccp{1.8cm}}
\toprule
\textbf{Model} & \textbf{EM (raw)} & \textbf{F1 (raw)} & \textbf{EM (norm)} & \textbf{F1 (norm)} & \textbf{Time (ms)} \\
\midrule
xlm\_roberta\_large & 0.992 & 0.994 & 0.992 & 0.994 & 22.4 \\
rubert\_base & 0.402 & 0.684 & 1.000 & 1.000 & 6.6 \\
rubert\_large & 0.398 & 0.679 & 1.000 & 1.000 & 6.5 \\
xlm\_roberta\_base (w/o fine-tun.) & 0.440 & 0.574 & -- & -- & 22.4 \\
rubert\_base (w/o fine-tun.) & 0.068 & 0.187 & -- & -- & 6.6 \\
rule\_based & 0.492 & 0.492 & -- & -- & \(\approx\)0 \\
t5\_rus\_base (w/o fine-tun.) & 0.176 & 0.220 & -- & -- & 27.2 \\
\end{tabular}
\caption{QA model results on the validation set (raw and normalized metrics)}
\label{tab:qa_results}
\end{table}

\begin{figure}[htbp]
\centering
\includegraphics[width=0.8\linewidth]{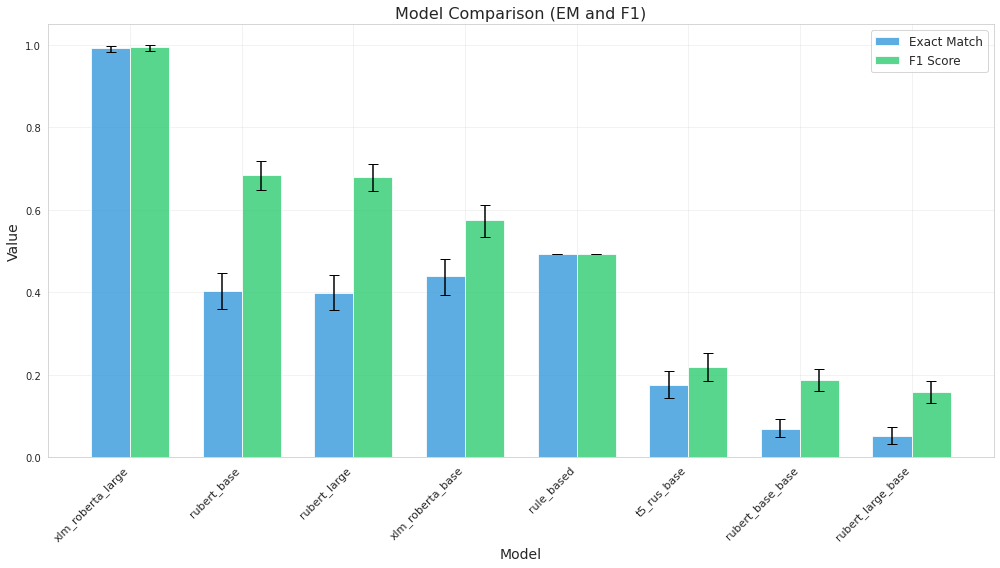}
\caption{Comparison of models by Exact Match and F1 score after normalization (error bars represent 95\% confidence intervals).}
\label{fig:qa_em_f1}
\end{figure}

The multilingual XLM-RoBERTa large initially delivers virtually perfect answers (\(\text{EM} = 0.992\)) and requires no post-processing. Both RuBERT models, after minimal normalization, achieve exact match with the reference (\(\text{EM} = 1.000\)), while operating 3.5 times faster---6.5 ms versus 22.4 ms per query. This speed advantage makes RuBERT the preferred choice for production systems, provided a thin post-processing layer is added. Figure~\ref{fig:qa_em_f1} presents a comparison of models in terms of EM and F1 with 95\% bootstrap confidence intervals, providing a visual complement to the tabular data.

While the overall metrics presented in Figure~\ref{fig:qa_em_f1} are encouraging, a nuanced understanding of the models' capabilities requires a deeper dive into their performance across different question categories. To this end, we conducted a fine-grained per-category quality analysis, presented in Table~\ref{tab:qa_per_category}. After normalization, all three fine-tuned models achieve 100\% F1 on questions about etymology, location, region, and object type. The only noticeable deviation is observed for XLM-RoBERTa large on coordinates (\(\text{F1} = 0.984\)), which is associated with rare cases of inaccurate copying of the last digit of a decimal fraction. This minor degradation does not affect the RuBERT models, which achieve perfect scores across all categories after post-processing.

\begin{table}[htbp]
\centering
\small
\begin{tabular}{lcccccc}
\toprule
\textbf{Model} & \textbf{Coordinates} & \textbf{Etymology} & \textbf{Location} & \textbf{Region} & \textbf{Sources} & \textbf{Object Type} \\
\midrule
xlm\_roberta\_large & 0.984 & 1.000 & 1.000 & 1.000 & 0.988 & 1.000 \\
rubert\_base & 1.000 & 1.000 & 1.000 & 1.000 & 1.000 & 1.000 \\
rubert\_large & 1.000 & 1.000 & 1.000 & 1.000 & 1.000 & 1.000 \\
\end{tabular}
\caption{F1 score by question type after normalization}
\label{tab:qa_per_category}
\end{table}

\begin{figure}[htbp]
\centering
\includegraphics[width=0.8\linewidth]{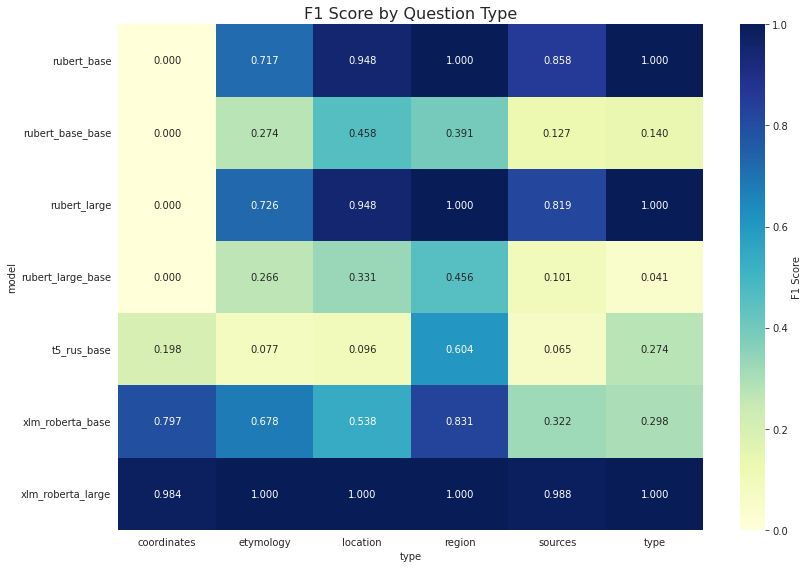}
\caption{Heatmap of F1 score by question type for fine-tuned models (after normalization).}
\label{fig:qa_heatmap}
\end{figure}

To provide an intuitive and comprehensive visualization of this per-category analysis, we present a heatmap in Figure~\ref{fig:qa_heatmap}. This representation immediately reveals the overall structure of model performance: all cells are coloured in the warmest tones, corresponding to an F1 of unity, with the single exception of the coordinates category for XLM-RoBERTa-large. The lighter cell visually confirms the minor but consistent deviation reported in Table~\ref{tab:qa_per_category}. The heatmap not only facilitates rapid cross-model and cross-category comparison but also highlights the structural similarity of performance across RuBERT-based models, which exhibit uniformly perfect scores. The stark contrast between the uniformly dark rows for RuBERT and the slightly lighter cell for XLM-RoBERTa-large underscores the effectiveness of the post-processing strategy when applied to monolingual architectures.

The detailed per-category breakdown and its heatmap visualization confirm that, following the appropriate normalization, the RuBERT models deliver flawless performance across every question type. While these results validate the model's theoretical capability, they do not fully address the practical constraints of real-world deployment. Two additional dimensions are critical for assessing the readiness of the system for production use: the structural consistency of generated outputs and the computational efficiency of inference. The first dimension ensures that the model produces answers in a format that can be seamlessly integrated into downstream applications, while the second determines the feasibility of deploying the system in latency-sensitive environments such as real-time geospatial services or interactive digital archives. The final part of the experimental study addresses these aspects through a systematic analysis of answer length distributions and computational latency, providing a comprehensive assessment of the system's practical deployability.

\begin{figure}[htbp]
\centering
\includegraphics[width=0.8\linewidth]{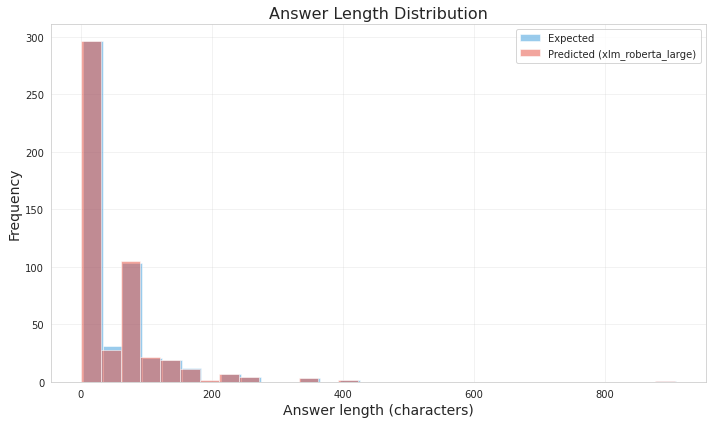}
\caption{Distribution of answer length (in characters) for reference values and predictions of the xlm\_roberta\_large model.}
\label{fig:answer_length}
\end{figure}

The distribution of predicted answer lengths for the XLM-RoBERTa large model is practically identical to the reference, as shown in Figure~\ref{fig:answer_length}. The bulk of answers is concentrated in the 20--40 character range, corresponding to coordinates and short names; there is also a small peak around 200 characters, corresponding to bibliographic sources. For RuBERT after normalization, the distribution shape coincides with the reference without significant deviations, indicating that the post-processing step does not distort the natural answer length distribution.

\begin{figure}[htbp]
\centering
\includegraphics[width=0.8\linewidth]{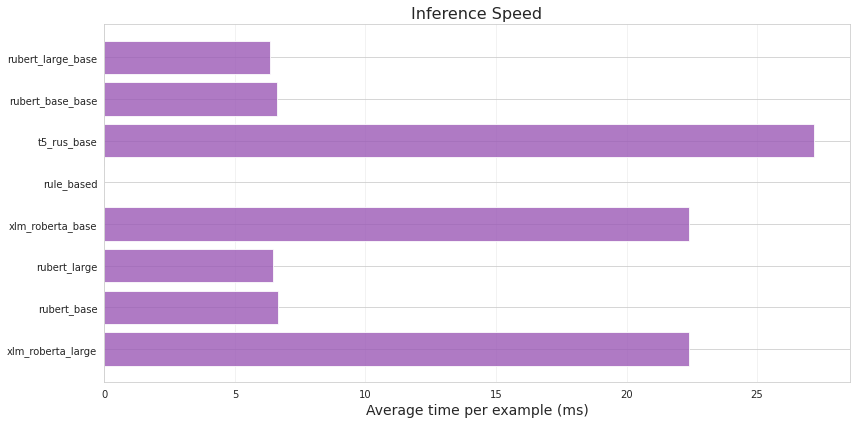}
\caption{Average processing time per example (ms) for the considered QA models.}
\label{fig:inference_time}
\end{figure}

While the high-quality answer extraction demonstrated in Figure~\ref{fig:answer_length} is encouraging, the practical applicability of the proposed system in real-world scenarios depends not only on accuracy but also on computational efficiency. To fully assess the trade-offs between precision and latency, we performed a detailed benchmark of inference times across all evaluated models. Speed measurements (Figure~\ref{fig:inference_time}) show that RuBERT spends only about 6.5 ms per query, XLM-RoBERTa large spends 22.4 ms, and the generative T5 spends 27.2 ms. Thus, for tasks where latency is critical, RuBERT with post-processing appears to be the optimal alternative, whereas XLM-RoBERTa large may be preferable when maximum quality of numerical coordinate extraction is required without additional programming.

The contrast between the two figures is particularly informative. Figure~\ref{fig:answer_length} confirms that all models correctly capture the structural characteristics of the target answers, regardless of their length. This means that the accuracy gains obtained through fine-tuning are not achieved at the expense of generating malformed or structurally inconsistent outputs. The models learn not only the content but also the natural distribution of answer lengths. In contrast, Figure~\ref{fig:inference_time} reveals a substantial divergence in computational overhead across architectures. The 3.5× speed advantage of RuBERT over XLM-RoBERTa-large, combined with its perfect accuracy after normalization, makes it a compelling choice for high-throughput applications. This finding challenges the common assumption that multilingual models are universally superior for multilingual tasks. In scenarios where post-processing is permissible, a carefully fine-tuned monolingual model may offer both superior speed and comparable accuracy.

In summary, these inference speed measurements provide a clear practical trade-off for system designers. The near-instantaneous inference of RuBERT, combined with simple post-processing, offers a highly efficient solution for large-scale or real-time applications, whereas the robust native output of XLM-RoBERTa-large is advantageous for tasks demanding maximum precision with minimal post-processing overhead.

Thus, the totality of the experimental data convincingly demonstrates that the proposed two-component architecture achieves near-ideal performance at all stages---from retrieval of the relevant document to precise extraction of the answer---and the identified tokenization peculiarities of RuBERT not only explain previous observations but also provide a simple recipe for their complete elimination.

\section{Discussion}
The experimental results presented above allow us to draw several interconnected conclusions about the developed system and, more broadly, about building question-answering pipelines over structured bilingual toponymic data. The most significant finding is that a lightweight hybrid architecture, coupling a state-of-the-art multilingual dense retriever with a straightforward spatial filter, achieves practically ideal recall for geographic queries at the scale of a single region. The fact that the optimal contribution of the semantic component turned out to be as low as \(\alpha = 0.1\) quantitatively confirms an intuitive but previously unmeasured insight: for toponym search, geographic proximity is the dominant relevance signal. Once the search space is narrowed by a 50 km radius, even a modest semantic signal suffices to pinpoint the correct object.

This observation has direct consequences for the design of geospatial retrieval systems in low-resource multilingual settings. Rather than investing heavily in complex neural rankers, one can achieve near-perfect performance by concentrating on data quality—accurate coordinates, rich contextual descriptions, and a mechanism for resolving homonymy. The manual inspection of the six retrieval errors, which were all traced to coordinate inaccuracies, namesakes, or sparse contexts (see Section~5), reinforces this message: the algorithm itself is robust; further gains lie primarily in data curation.

The extractive reading experiments revealed an equally instructive, and somewhat counterintuitive, pattern. The apparent failure of monolingual RuBERT models on coordinate questions was not due to any semantic deficiency but solely to WordPiece tokenization artifacts that inserted spurious spaces into numerical strings. Trivial post-processing completely eliminated the issue, allowing both RuBERT variants to reach perfect accuracy while being 3.5 times faster than the multilingual XLM-RoBERTa-large. This result challenges the widespread assumption that multilingual SentencePiece models are inherently necessary for handling numerical data. In practical deployments where a thin post-processing layer is acceptable, monolingual models can offer a superior speed–accuracy trade-off—a valuable lesson for NLP engineering in low-resource language contexts.

The high performance of the individual components suggests that an end-to-end RAG pipeline can be built with confidence, as the retrieval module guarantees that the correct context is supplied to the reader in virtually every case. Although a full end-to-end evaluation remains to be conducted, the near-ceiling metrics at both stages provide a strong upper-bound estimate of the overall system quality.

Several limitations must be kept in mind. The QA corpus and the test queries were generated from templates, which, while guaranteeing answer localization, do not fully reflect the variability of real user queries. Expanding the evaluation to include naturalistic questions is an important next step. The geospatial filter uses a fixed 50 km radius, which may be suboptimal for objects of very different spatial scales (e.g., a mountain range vs. a spring). Furthermore, physiographic characteristics constituted only 1.6\% of the QA corpus, so conclusions about this category remain preliminary.

The open release of all resources—the bilingual toponymic dataset, the QA corpus, the fine-tuned models, and the web demonstrator—ensures reproducibility and invites follow-up work. The inclusion of etymological information, questioned by one reviewer, proves valuable for explanatory question answering, which is increasingly demanded by educational platforms, digital tourism services, and cultural heritage projects. In such applications, the ability to answer not only \textit{where} but also \textit{why} a place is named a certain way adds a qualitatively new dimension to geospatial exploration.

Future work will pursue adaptive radius selection, coordinate cross-validation against external sources (e.g., OpenStreetMap), and a full end-to-end RAG evaluation including generative models. The proposed methodology is portable to other multilingual regions, provided that toponymic datasets of comparable structure and coordinate coverage can be constructed.

\section{Conclusion}
This work has demonstrated that a carefully designed, yet conceptually simple, hybrid question-answering system can achieve near-perfect performance on bilingual geospatial queries over structured toponymic data. By combining multilingual dense embeddings with lightweight spatial filtering, and by pairing the resulting retriever with fine-tuned extractive readers, we obtained an end-to-end pipeline that locates the correct document and pinpoints the exact answer in almost every case. The key practical insights—the dominance of geographic proximity for toponym search and the speed advantage of monolingual models after tokenization-aware post-processing—extend beyond the immediate setting and can inform the design of similar systems for other low-resource languages.

The main scientific and practical contributions are as follows:

\begin{enumerate}
\item An openly available bilingual (Russian–Tatar) dataset of 9,688 toponyms with rich linguistic, etymological, and coordinate information, together with a specialized extractive QA corpus of 38,696 span-annotated pairs, was created and published \citep{tatarnlpworld_toponyms, tatarnlpworld_toponyms_qa}.

\item A hybrid retrieval method integrating dense semantic indexing with geospatial ranking was shown to achieve \(\text{Recall@5} = 1.000\) and \(\text{MRR} = 0.994\), significantly outperforming BM25, purely semantic, and purely spatial alternatives.

\item A multi-architecture benchmark of extractive readers demonstrated that XLM-RoBERTa-large reaches \(\text{EM} = 0.992\) without post-processing, while RuBERT models, after trivial normalization that compensates for WordPiece tokenization artifacts, achieve perfect accuracy with 3.5 times faster inference.

\item All resources—dataset, QA corpus, model weights, and an interactive web demo—are released on Hugging Face, guaranteeing full reproducibility and immediate applicability.
\end{enumerate}

The developed system is ready for integration into geoinformation services, digital archives, educational platforms, and cultural heritage preservation projects. The near-perfect retrieval recall ensures that downstream generative models, when incorporated into a full RAG pipeline, will receive relevant context in virtually all cases, effectively eliminating hallucination risks originating from the retrieval stage.

Further research will address adaptive spatial filtering, coordinate verification against external sources, end-to-end RAG evaluation, and the expansion of the QA corpus with open-ended questions. The approach can be scaled to other bilingual regions, provided that toponymic datasets with comparable structure and coordinate referencing are available.

\bibliographystyle{unsrtnat}
\bibliography{references}

@article{suleymanov2021_system,
  author  = {Suleymanov, D. Sh. and Fridman, A. Ya. and Gilmullin, R. A. and Kulik, B. A.},
  title   = {System Analysis of the Problem of Natural Language Modeling},
  journal = {Transactions of the Kola Science Centre. Information Technologies},
  year    = {2021},
  volume  = {12},
  number  = {5},
  pages   = {57--66},
  note    = {(In Russian)}
}

@inproceedings{galimov2023_dialectologist,
  author    = {Galimov, M. and Burnashev, R. and Gatiatullin, A.},
  title     = {Designing a Prototype of a Fuzzy Expert System for a Dialectologist Using Geographic Information Systems and Technologies},
  booktitle = {Proceedings of the 8th International Conference on Computer Science and Engineering (UBMK)},
  address   = {Burdur, Turkiye},
  year      = {2023},
  pages     = {382--386}
}

@inproceedings{rehurek2006_gensim,
  author    = {Rehurek, R. and Sojka, P.},
  title     = {Gensim -- Python Framework for Vector Space Modelling},
  booktitle = {Proceedings of the 9th International Conference on Text, Speech and Dialogue (TSD)},
  address   = {Brno, Czech Republic},
  year      = {2006},
  pages     = {45--50}
}

@misc{saykhunov_corpus_tatar,
  author  = {Saykhunov, M. R. and Khusainov, R. R. and Ibragimov, T. I.},
  title   = {Written Corpus of the Tatar Language},
  year    = {2025},
  url     = {https://www.corpus.tatar/en},
  note    = {Accessed: 03.01.2026}
}

@misc{sketch_engine_tatar,
  author  = {{Sketch Engine}},
  title   = {Tatar Mixed Corpus -- Tatar Corpus from the Web},
  year    = {2024},
  url     = {https://www.sketchengine.eu/tatar-corpus-from-the-web/},
  note    = {Accessed: 03.01.2026}
}

@misc{ipsan_tat_monocorpus,
  author  = {{IPSAN}},
  title   = {tat\_monocorpus\_v2},
  year    = {2024},
  url     = {https://huggingface.co/datasets/IPSAN/tat_monocorpus_v2/},
  note    = {Accessed: 28.02.2026}
}

@inproceedings{gilmullin2017_morphological,
  author    = {Gilmullin, R. A. and Gataullin, R. R.},
  title     = {Morphological Analysis System of the Tatar Language},
  booktitle = {Proceedings of the 9th International Conference on Computational Collective Intelligence (ICCCI)},
  publisher = {Springer, Cham},
  year      = {2017},
  pages     = {519--528}
}

@misc{arabov2026_tatartokenizers,
  author  = {Arabov, M. K.},
  title   = {TatarTokenizers},
  year    = {2026},
  note    = {Certificate of State Registration of Computer Program No. 2026611049 dated 16.01.2026, Kazan (Volga Region) Federal University, Kazan. (In Russian)}
}

@inproceedings{mukhamedshin2025_semantic,
  author    = {Mukhamedshin, D. R. and Gatiatullin, A. R. and Prokopyev, N. A. and Gilmullin, R. A.},
  title     = {Semantic Annotation in Electronic Corpus of Tatar Language `Tugan Tel' Based on Knowledge Graph},
  booktitle = {Proceedings of the 10th International Conference on Computer Science and Engineering (UBMK)},
  address   = {Istanbul, Turkiye},
  year      = {2025},
  pages     = {1792--1795}
}

@inproceedings{mindubaev2024_semantic_relation,
  author    = {Mindubaev, A. and Gatiatullin, A. R.},
  title     = {Problems of Semantic Relation Extraction from Tatar Text Corpus `Tugan Tel'},
  booktitle = {Proceedings of the 3rd International Conference on Problems of Informatics, Electronics and Radio Engineering (PIERE)},
  address   = {Novosibirsk, Russian Federation},
  year      = {2024},
  pages     = {1690--1694}
}

@misc{arabov2026_tatar2vec,
  author  = {Arabov, M. K.},
  title   = {Tatar2Vec},
  year    = {2026},
  note    = {Certificate of State Registration of Computer Program No. 2026610619 dated 14.01.2026, Kazan (Volga Region) Federal University, Kazan. (In Russian)}
}

@inproceedings{gafarov2025_explainable,
  author    = {Gafarov, F. M. and Gafarova, V. R. and Ayupov, M. M.},
  title     = {Explainable Artificial Intelligence Methods in Text Classification of Machine Learning Models for Tatar Language},
  booktitle = {Proceedings of the 10th International Conference on Computer Science and Engineering (UBMK)},
  address   = {Istanbul, Turkiye},
  year      = {2025},
  pages     = {1796--1800}
}

@misc{liang2022_e5,
  author  = {Liang, P. and Zhang, M. and others},
  title   = {EmbEddings from bidirEctional Encoder rEpresentations (E5): A Universal Embedding Model for Information Retrieval},
  year    = {2022},
  url     = {https://arxiv.org/abs/2212.03533},
  note    = {arXiv:2212.03533. Accessed: 28.02.2026}
}

@inproceedings{conneau2020_xlmr,
  author    = {Conneau, A. and Khandelwal, K. and Goyal, N. and Chaudhary, V. and Wenzek, G. and Guzm{\'a}n, F. and Grave, E. and Ott, M. and Zettlemoyer, L. and Stoyanov, V.},
  title     = {Unsupervised Cross-lingual Representation Learning at Scale},
  booktitle = {Proceedings of the 58th Annual Meeting of the Association for Computational Linguistics (ACL)},
  address   = {Online},
  year      = {2020},
  pages     = {8440--8451}
}

@misc{geonames,
  author  = {{GeoNames}},
  title   = {GeoNames},
  year    = {2025},
  url     = {https://www.geonames.org/},
  note    = {Accessed: 28.02.2026}
}

@misc{tatarnlpworld_hf,
  author  = {{TatarNLPWorld}},
  title   = {Turkic NLP \& Low Resource Languages Research Hub},
  year    = {2025},
  url     = {https://huggingface.co/TatarNLPWorld},
  note    = {Accessed: 28.02.2026}
}

@inproceedings{schweter2020_berturk,
  author    = {Schweter, S. and Tekir, S.},
  title     = {BERT for Turkish: A Comprehensive Evaluation},
  booktitle = {Proceedings of the 12th Language Resources and Evaluation Conference (LREC)},
  address   = {Marseille, France},
  year      = {2020},
  pages     = {4325--4332}
}

@article{bentley1975_kdtree,
  author  = {Bentley, J. L.},
  title   = {Multidimensional Binary Search Trees Used for Associative Searching},
  journal = {Communications of the ACM},
  year    = {1975},
  volume  = {18},
  number  = {9},
  pages   = {509--517}
}

@article{sinnott1984_haversine,
  author  = {Sinnott, R. W.},
  title   = {Virtues of the Haversine},
  journal = {Sky and Telescope},
  year    = {1984},
  volume  = {68},
  number  = {2},
  pages   = {159}
}

@inproceedings{burnashev2025_geolinguistic,
  author    = {Burnashev, R. and Gatiatullin, A. and Khamidullin, M.},
  title     = {Geolinguistic System for Dialect Similarity Analysis Based on Associative Rules and Fuzzy Logic},
  booktitle = {Proceedings of the 10th International Conference on Computer Science and Engineering (UBMK)},
  address   = {Istanbul, Turkiye},
  year      = {2025},
  pages     = {1782--1785}
}

@inproceedings{rajpurkar2016_squad,
  author    = {Rajpurkar, P. and Zhang, J. and Lopyrev, K. and Liang, P.},
  title     = {SQuAD: 100,000+ Questions for Machine Comprehension of Text},
  booktitle = {Proceedings of the 2016 Conference on Empirical Methods in Natural Language Processing (EMNLP)},
  address   = {Austin, Texas},
  year      = {2016},
  pages     = {2383--2392}
}

@inproceedings{efimov2020_sberquad,
  author    = {Efimov, S. and others},
  title     = {SberQuAD -- Russian Reading Comprehension Dataset},
  booktitle = {Proceedings of the 22nd Conference on Artificial Intelligence (DCAI)},
  year      = {2020}
}

@inproceedings{devlin2019_bert,
  author    = {Devlin, J. and Chang, M.-W. and Lee, K. and Toutanova, K.},
  title     = {BERT: Pre-training of Deep Bidirectional Transformers for Language Understanding},
  booktitle = {Proceedings of the 2019 Conference of the North American Chapter of the Association for Computational Linguistics: Human Language Technologies (NAACL-HLT)},
  address   = {Minneapolis, Minnesota},
  year      = {2019},
  pages     = {4171--4186}
}

@article{raffel2020_t5,
  author  = {Raffel, C. and Shazeer, N. and Roberts, A. and Lee, K. and Narang, S. and Matena, M. and Zhou, Y. and Li, W. and Liu, P. J.},
  title   = {Exploring the Limits of Transfer Learning with a Unified Text-to-Text Transformer},
  journal = {Journal of Machine Learning Research},
  year    = {2020},
  volume  = {21},
  pages   = {1--67}
}

@inproceedings{choure2022_hindi_ner,
  author    = {Choure, A. A. and Adhao, R. B. and Pachghare, V. K.},
  title     = {NER in Hindi Language Using Transformer Model: XLM-Roberta},
  booktitle = {2022 IEEE International Conference on Blockchain and Distributed Systems Security (ICBDS)},
  address   = {Pune, India},
  year      = {2022},
  pages     = {1--5},
  doi       = {10.1109/ICBDS53701.2022.9935841}
}

@article{johnson2019_faiss,
  author  = {Johnson, J. and Douze, M. and J{\'e}gou, H.},
  title   = {Billion-scale Similarity Search with GPUs},
  journal = {IEEE Transactions on Big Data},
  year    = {2019},
  volume  = {7},
  number  = {3},
  pages   = {535--547}
}

@inproceedings{fenogenova2021_rusuperglue,
  author    = {Fenogenova, A. and Tikhonova, M. and Mikhailov, V. and Shavrina, T. and Emelyanov, A. and Shevelev, D. and Kukushkin, A. and Malykh, V. and Artemova, E.},
  title     = {Russian SuperGLUE 1.1: Revising the Lessons not Learned by Russian NLP models},
  booktitle = {Computational Linguistics and Intellectual Technologies: Papers from the Annual International Conference "Dialogue"},
  year      = {2021},
  number    = {20}
}

@inproceedings{artemova2022_runne,
  author    = {Artemova, E. and Zmeev, M. and Loukachevitch, N. and Rozhkov, I. and Batura, T. and Ivanov, V. and Tutubalina, E.},
  title     = {RuNNE-2022 Shared Task: Recognizing Nested Named Entities},
  booktitle = {Proceedings of the Dialogue 2022 Conference},
  address   = {Moscow},
  publisher = {RSUH},
  year      = {2022}
}

@misc{tatarnlpworld_toponyms,
  author  = {{TatarNLPWorld}},
  title   = {Tatarstan Toponyms Dataset},
  year    = {2025},
  url     = {https://huggingface.co/datasets/TatarNLPWorld/tatarstan-toponyms},
  note    = {Accessed: 28.02.2026}
}

@misc{tatarnlpworld_toponyms_qa,
  author  = {{TatarNLPWorld}},
  title   = {Tatarstan Toponyms QA Dataset},
  year    = {2025},
  url     = {https://huggingface.co/datasets/TatarNLPWorld/tatarstan-toponyms-qa},
  note    = {Accessed: 28.02.2026}
}

@book{garipova1984_hydronyms1,
  author    = {Garipova, F. G.},
  title     = {Tatarstan gidronimnarı süzlege. Berenche kitap},
  publisher = {Tatarstan kitap näşriyatı},
  address   = {Kazan},
  year      = {1984},
  note      = {(In Tatar)}
}

@book{garipova1990_hydronyms2,
  author    = {Garipova, F. G.},
  title     = {Tatarstan gidronimnarı süzlege. Ikenche kitap},
  publisher = {Tatarstan kitap näşriyatı},
  address   = {Kazan},
  year      = {1990},
  note      = {(In Tatar)}
}

@book{garipova1991_hydronymy,
  author    = {Garipova, F. G.},
  title     = {Issledovaniya po gidronimii Tatarstana},
  publisher = {Nauka},
  address   = {Moscow},
  year      = {1991},
  note      = {(In Russian)}
}

@book{garipova2010_toponyms,
  author    = {Garipova, F. G.},
  title     = {Tatar toponimnarı süzlege},
  address   = {Kazan},
  year      = {2010},
  note      = {(In Tatar)}
}

@book{sattarov1998_toponymy,
  author    = {Sattarov, G. F.},
  title     = {Tatar toponimiyase},
  publisher = {Kazan University Press},
  address   = {Kazan},
  year      = {1998},
  note      = {(In Tatar)}
}

@book{akhmetianov2015_etymology,
  author    = {Äkhmätÿanov, R. G.},
  title     = {Tatar telenen etimologik süzlege: Ike tomda},
  publisher = {Mägarif -- Vakıt},
  address   = {Kazan},
  year      = {2015},
  note      = {(In Tatar)}
}

@book{tatar_encyclopedia,
  editor    = {Khasanov, M. Kh.},
  title     = {Tatar ensiklopediyase: 6 tomda},
  publisher = {Institute of the Tatar Encyclopedia, Academy of Sciences of the Republic of Tatarstan},
  address   = {Kazan},
  year      = {2002--2014},
  note      = {(In Russian)}
}

@book{bayazitova2009_dialect,
  author    = {Bayazitova, F. S. and Ramazanova, D. B. and Sadykova, Z. R. and Khairetdinova, T. Kh.},
  title     = {Tatar telenen zur dialektologik süzlege},
  publisher = {Tatarstan kitap näşriyatı},
  address   = {Kazan},
  year      = {2009},
  note      = {(In Tatar)}
}

\end{document}